\documentclass{article}

\usepackage[preprint]{custom}
\usepackage[bookmarks=false]{hyperref}

\usepackage[utf8]{inputenc} 
\usepackage[T1]{fontenc}    
\usepackage{hyperref}       
\usepackage{url}            
\usepackage{booktabs}       
\usepackage{amsfonts}       
\usepackage{nicefrac}       
\usepackage{microtype}      
\usepackage{xcolor}         
\usepackage{amsmath}
\usepackage{graphicx}
\usepackage{algorithm}
\usepackage{algpseudocode}
\usepackage{tabularx}
\usepackage{subcaption} 
\usepackage{amssymb}

\usepackage{listings}

\newcommand{\Jinv}{\(\mathcal{J}_{\text{invalid}}\)}
\newcommand{\Jcur}{\(\mathcal{J}_{\text{current}}\)}
\newcommand{\Jval}{\(\mathcal{J}_{\text{valid}}\)}
\newcommand{\SCA}{\(\mathcal{A}_{\text{sv}}\)}

\title{HySem: A context length optimized LLM pipeline for unstructured tabular extraction}

\author{%
  Narayanan PP \thanks{Indian Institute of Science, Bangalore} \\
  \texttt{narayananp@iisc.ac.in} \\ \And
  Anantharaman Palacode Narayana Iyer \thanks{JNResearch Labs LLP, Bangalore} \\
  \texttt{ananth@jnresearch.com}
}

\begin{document}

\maketitle

\begin{abstract}
Regulatory compliance reporting in the pharmaceutical industry relies on detailed tables, but these are often under-utilized beyond compliance due to their unstructured format and arbitrary content. Extracting and semantically representing tabular data is challenging due to diverse table presentations. Large Language Models (LLMs) demonstrate substantial potential for semantic representation, yet they encounter challenges related to accuracy and context size limitations, which are crucial considerations for the industry applications. We introduce HySem, a pipeline that employs a novel context length optimization technique to generate accurate semantic JSON representations from HTML tables. This approach utilizes a custom fine-tuned model specifically designed for cost- and privacy-sensitive small and medium pharmaceutical enterprises. Running on commodity hardware and leveraging open-source models, HySem surpasses its peer open-source models in accuracy and provides competitive performance when benchmarked against OpenAI GPT-4o and effectively addresses context length limitations, which is a crucial factor for supporting larger tables.
\end{abstract}

\section{Introduction}
Data tables are essential for facilitating regulatory compliance and financial reporting across industries such as pharmaceuticals and finance. In the pharmaceutical sector, documents like Annual Product Quality Reviews (APQRs) often contain complex, ad-hoc structured tables that present significant challenges for integration into standard databases. While these tables hold valuable information, their unstructured format renders them non-queryable, hindering automated processing. Our objective is to convert these table-based assets, often stored in HTML format, into semantic JSON, which enables direct mapping of the generated JSON keys to the field names of the database schema, facilitating efficient integration and use in business analytics applications.

Processing real-world tables, particularly those prevalent in industries like pharmaceuticals, is a highly challenging task. Unlike structured databases with predictable schemas, pharmaceutical tables often have arbitrary placements of headers and data elements. Headers might appear mid-table, or there may be multiple, inconsistent levels of headers across different documents. \textbf{Appendix A.2} Figure \ref{fig:first-image} illustrates a complex pharmaceutical data table. 

The arbitrariness in table presentation renders rule-based approaches ineffective for transforming HTML tables into semantic JSON. Furthermore, developing and maintaining such algorithms is costly and does not scale efficiently to handle frequent changes in table formats. In contrast, large language models (LLMs) can be trained to recognize patterns and relationships within the data, offering a more robust, scalable and adaptable solution. However, LLMs often encounter performance challenges when handling complex tabular structures, particularly very long tables that contain multiple instances of pharma-specific terminology.

In this paper, we propose HySem, a pipeline designed to accurately transform HTML tables into semantic JSON representations (refer \textbf{Appendix A.3} for detailed illustrations of semantic JSON). Central to our approach is a novel Context Optimiser, which employs a dynamic token pruning technique to rewrite the input HTML tables. This process reduces the token count while maintaining the semantic integrity of the data. As a result, HySem can efficiently manage large and complex tables, including those featuring specialized pharmaceutical terminology, without sacrificing performance.
\\
Our research was motivated by the business needs of small and medium pharmaceutical enterprises, where cost and data privacy considerations are crucial. Specifically, our key goals are:
\\
\begin{itemize} \item \textbf{On-Premise Model Deployment:} Ensuring all models operate on-premise to maintain stringent data security standards. \item \textbf{Compatibility with Commodity Hardware:} Enabling models to run efficiently on commodity hardware with cost-effective, readily available GPUs. \item \textbf{Model Size Constraints:} Limiting model size to under 10 billion parameters, making it suitable for GPUs with up to 16 GB of memory. \item \textbf{Open-Source Software Stack:} Leveraging open-source tools and models to minimize costs and foster accessibility. \end{itemize}

Building on these foundational goals, we present the following contributions:
\\
\begin{itemize} \item \textbf{Context Optimizer:} A novel component that efficiently rewrites input data, significantly reducing token counts and enhancing processing speed. \item \textbf{Semantic Synthesizer:} A custom fine-tuned model designed to produce precise semantic representations from complex HTML tables. \item \textbf{Syntax Corrector:} An agentic system that identifies and corrects syntax errors in the output JSON with minimal human oversight. \item \textbf{Evaluation Methodology:} A new framework that evaluates the performance of our approach using a combination of intrinsic and extrinsic metrics. \end{itemize}


\section{Methodology}
Our proposed LLM Pipeline, HySem, aims to address the challenge of converting unprocessed raw HTML tables into a semantic JSON data structure while addressing the limitations of LLMs concerning context length and inference time. By adopting a novel strategy for optimizing the number of tokens, we aim to meet the industry standards for high accuracy and reduced inference time, particularly for regulatory compliance reporting. 

\begin{figure*}
    \centering
    \includegraphics[width=0.9\linewidth]{  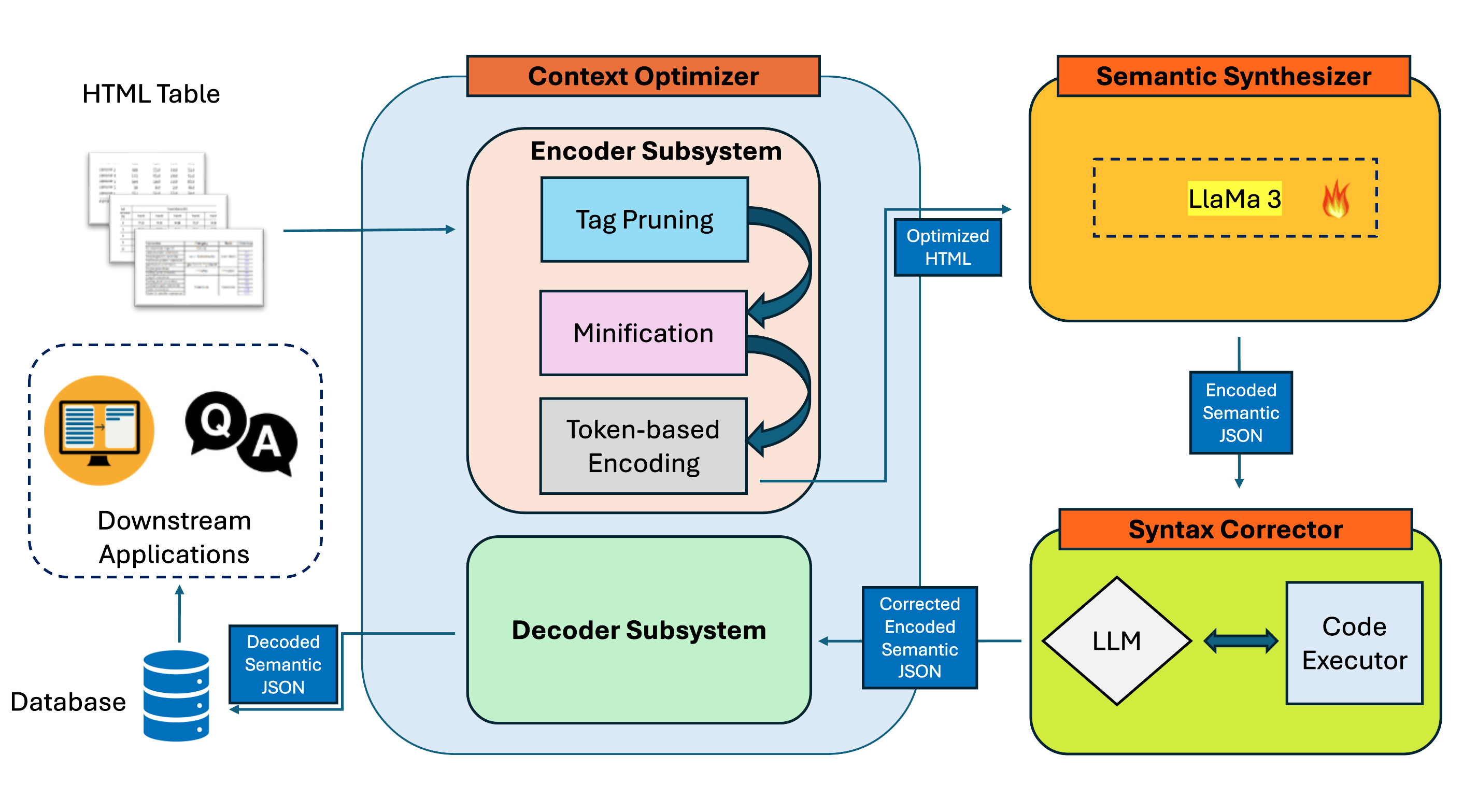}
    \caption{Hysem Architecture diagram}
    \label{fig:hysem-archi}
\end{figure*}

HySem achieves these requirements through a structured pipeline composed of three components: Context Optimizer (\( \mathcal{A}_{\text{CO}} \)), 
Semantic Synthesizer (\( \mathcal{A}_{\text{SS}} \)), Syntax Corrector (\( \mathcal{A}_{\text{SC}} \)). \( \mathcal{A}_{\text{CO}} \) employs a novel methodology for optimizing the context window utilized by HTML tables, enabling processing of large tables. \( \mathcal{A}_{\text{SS}} \) transforms this optimized HTML table into semantic JSON. \( \mathcal{A}_{\text{SC}} \) reviews the generated JSON for any syntax errors and outputs a syntactically-valid JSON. Figure \ref{fig:hysem-archi} illustrates the Hysem architecture in detail.

\subsection{Context Optimizer Subsystem}
The limited context size of large language models significantly constrains their ability to effectively process large tables \citep{liu2024lost, xu2024retrieval}. Although models with extended context lengths can alleviate this issue, they may sacrifice accuracy and increase processing times due to the quadratic complexity of self-attention mechanisms \citep{sui2024table, Tay2020EfficientTA, vaswani2017attention}. Recent advancements in Transformer architectures, such as Linformer \citep{wang2020linformer} and Flash Attention techniques \citep{dao2022flashattention}, have shown improvements in memory and time complexity.  Nevertheless, the critical need to reduce token count persists, as it is essential for enabling faster inference times and optimizing the utilization of limited GPU memory, all without sacrificing model performance.

Our approach to optimizing context length utilization stems from the observation that significant token inefficiency arises when there is a mismatch between the tokenizer's vocabulary and domain-specific terminology in the input text. For example, "Amoxycillin," a widely recognized pharmaceutical medication, is not present in the Llama 3 tokenizer's vocabulary, leading to its representation by multiple tokens. In addition to domain-specific terminology, other data types commonly encountered in real-world tables, such as dates, string identifiers, and proper nouns, often fail to align with the tokenizer’s vocabulary, further increasing token count and exacerbating inefficiency.

We introduce \textbf{Context Optimizer}, a novel token alignment technique designed to minimize the impact of "token-vocabulary" misalignment. This approach optimizes token representation by aligning cell contents in a table with the tokenizer’s vocabulary, thereby reducing the number of tokens required to represent each cell. 

The Context Optimizer operates in 2 phases: \textit{Encoding phase} where we rewrite the cell contents into a more compact form and \textit{Decoding phase} which restores the original contents.

\subsubsection{Encoding Phase} 
In the encoding phase, our process is divided into two primary stages: standard pre-processing and a specialized token-based encoding method.

\textbf{Pre-processing Steps}

We begin with a set of conventional pre-processing steps. First, we remove tags and attributes from the HTML tables that do not contribute to semantic understanding, such as those meant for styling. After this, a minification step is applied, which strips away unnecessary white spaces, further optimizing the HTML table for encoding.

\begin{figure*}
    \centering
    \begin{subfigure}{0.95\linewidth}
        \centering
        \includegraphics[width=\linewidth, height=0.12\textheight]{  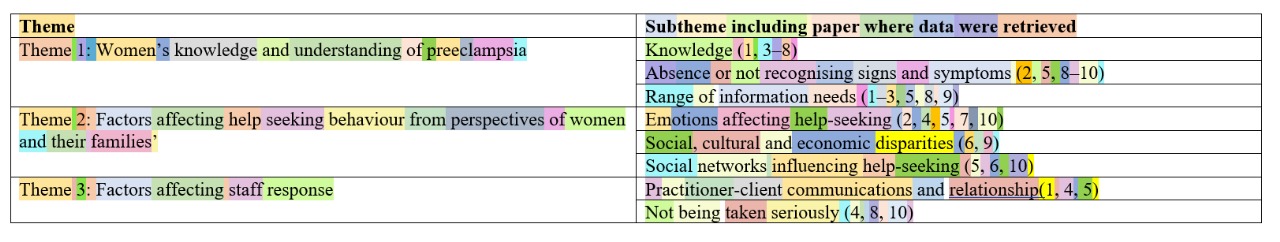}
        \caption{HTML Table before optimization}
        \label{fig:html-tab-bef-opt}
    \end{subfigure}
    
    \vspace{1em} 

    \begin{subfigure}{0.65\linewidth}
        \centering
        \includegraphics[width=\linewidth, height=0.12\textheight]{  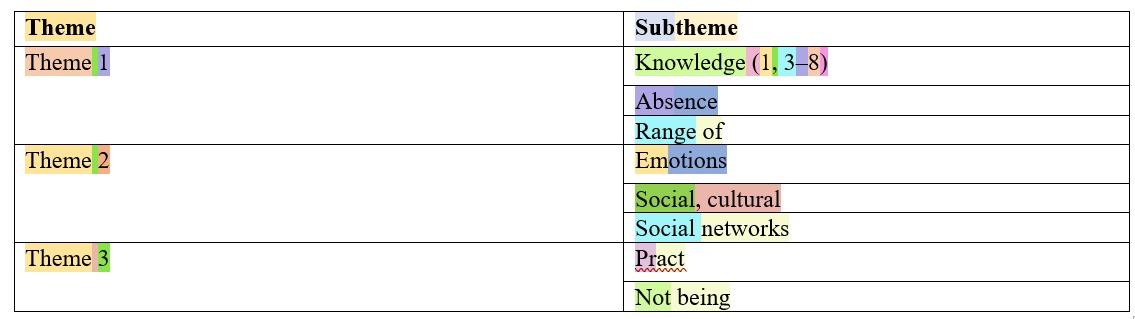}
        \caption{HTML Table after optimization}
        \label{fig:html-tab-aft-opt}
    \end{subfigure}
    
    \caption{An illustration of a table optimized by the Context Optimizer is shown in \ref{fig:html-tab-bef-opt}, presenting the original HTML table without any optimization. \ref{fig:html-tab-aft-opt} displays the same table after optimization. In these figures, tokens within each cell are highlighted with distinct colors to facilitate easy observation of the token count per cell. The tokenization is performed using the LLaMa 3 tokenizer.}
    \label{fig:hysem-both}
\end{figure*}

\textbf{Token-Based Encoding}

Next, we apply our custom token-based encoding technique. The goal here is to represent the content of each cell with the minimum number of tokens while ensuring that each cell has a unique representation. The algorithm for this process is detailed in Algorithm \ref{alg:token_enc_alg}.

Prior to encoding, we first sort the cells in ascending order based on the number of tokens they contain. This strategy allows us to resolve \textit{potential collisions} more easily, as cells with fewer tokens are processed first. A collision occurs when two distinct cell contents map to overlapping token sequences.


Our encoding process incorporates several high-level heuristics to enhance efficiency and accuracy:

\textbf{a. Single Token Preservation:} If a cell's content consists of a single token, it remains unchanged.

\textbf{b. Multi-Token Optimization:} For cells with multiple tokens, we aim to represent the content using only two tokens whenever possible. This approach addresses the initial attempt of encoding with a single token, which often compromised semantic richness and resulted in inaccurate JSON predictions. 

\textbf{c. Bracket Handling:} We handle incomplete bracket sequences by checking if a token starts with an opening bracket (e.g., [, \{) and lacks a matching closing bracket. In such cases, we concatenate subsequent tokens until the bracket is closed. This approach is crucial to prevent syntax errors in the generated JSON.

Figure \ref{fig:hysem-both} offers a detailed illustration of the HTML table before and after optimization. For example, in Figure \ref{fig:html-tab-bef-opt} the cell content \textit{"Theme 1: Women’s knowledge and understanding of preeclampsia"} is tokenized into 15 distinct tokens, represented by various colors. After optimization, the same cell content transforms into \textit{"Theme 1"}, consuming only 3 tokens as illustrated in Figure \ref{fig:html-tab-aft-opt}. 


The overall objective is to use the fewest tokens possible while maintaining uniqueness across all cell contents. By treating each tokenized cell as a unit, we can reduce the total number of tokens significantly without losing semantic integrity. The encoded HTML table is processed by the downstream modules in the HySem pipeline, including the Semantic Synthesizer and the Syntax Corrector.

\begin{algorithm}
\caption{Token-Based Encoding Algorithm}\label{alg:token_enc_alg}
\textbf{Input:} Table $T$ with $n$ cells $\{C_1, C_2, \dots, C_n\}$, each containing text. \\
\textbf{Output:} Encoded table $T' = \{E_1, E_2, \dots, E_n\}$ with unique token representations.

\begin{algorithmic}[a]
    \State Sort cells $\{C_1, C_2, \dots, C_n\}$ in ascending order based on the number of tokens in each cell.
    
    \For{each cell $C_i \in T$ (after sorting)}
        \State Tokenize $C_i$: $T_i = \{t_1, t_2, \dots, t_k\}$ where $t_j$ are tokens of $C_i$.
        \State Apply heuristics to determine the initial encoded form $E_i$.  \Comment{Use predefined strategies}
        \\
        \If{$\exists j < i$ such that $E_i = E_j$} \Comment{Check if a previous encoding $E_j$ is the same as $E_i$}
            \For{$t_k \in \{t_2, t_3, \dots, t_k\}$}
                \State $E_i = E_i \cup t_k$.  \Comment{Concatenate additional tokens}
                \If{$E_i \neq E_j$}
                    \State \textbf{break}.  \Comment{Stop once the encoding is unique}
                \EndIf
            \EndFor
        \EndIf
        
        \State Store the mapping $(C_i, E_i)$.
    \EndFor
    \State \textbf{return} encoded table $T' = \{E_1, E_2, \dots, E_n\}$.
\end{algorithmic}
\end{algorithm}


\subsubsection{Decoding Phase}

In the decoding phase, the output generated by the LLM is decoded to restore the original lexicon used in the table. This decoding step reverses the earlier encoding which is present in the model generated JSON, reconstructing the content to accurately reflect the contents of the original HTML table while preserving the benefits of token optimization. The resulting JSON is both semantically accurate and more efficiently processed.

A key feature of our Context Optimizer is its \textbf{dynamic nature}, where the mapping of input words to optimized token sequences, as well as the corresponding decoding process, are entirely driven by the current input. This process operates without reliance on any pre-defined static mappings.

\subsection{Semantic Synthesizer}
Given the LLM's capability to comprehend deep semantic relationships, HySem adopts the open-sourced Meta-Llama-3-8B-Base model and fine-tunes it with a manually labeled dataset for transforming HTML tables into semantic JSON.

Concretely, the Semantic Synthesizer \( \mathcal{A}_{\text{SS}} \) accepts a HTML table \( \mathcal{H}_{\text{i}} \) optimized by the Context Optimizer, as input and produces JSON \( \mathcal{J}_{\text{i}} \) as the output in the same encoded semantic space as the optimized input HTML table. Our initial experiments indicate that adopting Prompt Engineering to achieve this transformation results in the generation of less accurate JSON representations, as evidenced by our Intrinsic and Extrinsic evaluations. We found that the LLM is highly sensitive to prompts and struggles to effectively capture the wide variety present in these tables. 
We identified several common failure patterns present in the JSONs generated by the LLM, which are cataloged as \textit{Semantic Failures} in Table \ref{tab:semantic_failures}. 

To address these failure modes, we fine-tune the LLM to generate JSON representations that accurately reflect the input HTML table.

\textbf{Dataset}

For our dataset, we require inputs as HTML tables and labels as semantic JSON. We utilize the following two open-sourced datasets for tabular HTML sources:

\textbf{PubTabNet}: PubTabNet \citep{smock2022pubtables} is a large-scale dataset for image-based table recognition, containing over 568,000 images of tables from scientific papers along with their corresponding HTML annotations.

\textbf{FinTabNet}: FinTabNet \citep{zheng2021global} is a dataset specifically designed for recognizing tables in financial documents, containing over 112,000 tables. Each table instance annotation includes fields such as the HTML structure and bounding box coordinates, similar to PubTabNet.

For our purposes, we extract the HTML from these sources and hand-label the corresponding JSON annotations for these tables. We filter tables that fit within the LLM's context length (8k context window for Llama3) and select 1,364 HTML tables from both PubTabNet and FinTabNet. We split the data into 756 training samples and 608 testing samples. We chose these datasets as they are public and are reasonable representations of the Pharmaceutical and Finance verticals respectively.

We also created additional hand-labeled datasets to address the custom needs of the industry using the proprietary customer supplied data. These proprietary datasets are used to fine-tune models that use Semantic Synthesizer LLM as the base model.


\begin{table*}[h]
\centering
\begin{tabular}{p{0.3\linewidth} p{0.6\linewidth}} 
\toprule
\textbf{Failure Mode} & \textbf{Example} \\ 
\midrule
Lexical errors & "Characteristics" in place of "Characteristic" \\ 
Missing words & "123" instead of "123 (n=25)" \\ 
Missing entire rows & An entire row under a subheading missed \\ 
Misaligned keys/values & A cell value placed under a different parent key \\ 
Merged Cells & "95\% CI": ["- 211.5", "69.0"] $\Rightarrow$ "95\% CI": "- 211.5\textbackslash n69.0" \\ 
Unicode errors & "0.88 (0.53 \textbackslash u2013 1.33)" $\Rightarrow$ "0.88 (0.53 - 1.33)" \\ 
\bottomrule
\end{tabular}
\caption{Semantic Failures}
\label{tab:semantic_failures}

\end{table*}

\begin{table*}[h]
\centering
\begin{tabularx}{\textwidth}{l>{\raggedright\arraybackslash}X}
\toprule
\textbf{Failure Mode} & \textbf{Description} \\
\midrule
Missing List Enclosure & The output is expected to be a list of dictionaries, but the LLM generates dictionaries without enclosing them in a list. \\ 
Unmatched Curly Braces & The JSON output is missing one or more curly braces, resulting in an incomplete or unbalanced structure. \\ 
Missing Commas & The JSON output is missing commas, leading to improperly formatted data. \\
Incorrect Placement of Quotes & For long numeric strings, like 123,456,789, the output should enclose the entire string in quotes. Instead, the quotes are incorrectly inserted within the string, producing a format like 123,"456,789". \\ 
\bottomrule
\end{tabularx}
\caption{Structural Failures}
\label{tab:syntax_err}

\end{table*}

\subsection{Syntax Corrector}
Syntax errors in the LLM-generated JSON output render the table unusable for further processing, such as ingestion into databases. Consequently, correcting these syntax errors is a critical functionality, especially for enabling automated workflows in industrial settings. To address these challenges, we developed a Syntax Corrector, based on \textit{reflective agentic framework}. 

Specifically, the Syntax Corrector \( \mathcal{A}_{\text{sc}} \) accepts a syntactically invalid JSON \( \mathcal{J}_{\text{i}} \) as input and produces a syntactically valid JSON \( \mathcal{J}_{\text{v}} \) through \textit{iterative refinement}. Through self-reflection \citep{ji-etal-2023-towards, pan-etal-2024-automatically, madaan2024self, asai2023self, renze2024self}, \( \mathcal{A}_{\text{sc}} \) iteratively refines the JSON output until a syntactically valid result is achieved or the maximum number of iterations is reached. The algorithm for this process is detailed in Algorithm \ref{alg:syntax_corrector} of \textbf{Appendix A.5}. Table \ref{tab:syntax_err} displays the common syntax error patterns observed in the LLM output.

\section{Evaluation Methodology}
For the accurate transformation of HTML tables into semantic JSON, two key objectives must be satisfied. First, \textit{\textbf{content preservation:}} all strings from the HTML table cells must exactly occur in the JSON output, ensuring no information is lost during the conversion. Second, \textit{\textbf{semantic accuracy:}} the generated JSON must accurately represent the hierarchical and relational structure of the original HTML table. 

Building on this foundation, we develop intrinsic and extrinsic evaluation methods to measure the content and semantic accuracy of the JSON produced by HySem. \textbf{Appendix A.4} provides detailed illustrations for intrinsic and extrinsic evaluation methods.

\subsection{Intrinsic Evaluation}
In intrinsic evaluation, we assess the representation of content from HTML cells in the generated JSON. We parse the HTML using Beautiful Soup to extract the \textit{set of all the cell contents} present in the HTML input, denoted as \( \mathcal{H}_{\text{set}} = \{ c_1, c_2, \ldots, c_m \} \). Similarly, we take the \textit{set of all elements} in the JSON, denoted as \( \mathcal{J}_{\text{set}} = \{ e_1, e_2, \ldots, e_n \} \) 

To evaluate if each content item \( c_i \) from \( \mathcal{H}_{\text{set}} \) is present in \( \mathcal{J}_{\text{set}} \), we define an indicator function $I(c_i \in \mathcal{J}_{\text{set}})$:
\[
I(c_i \in \mathcal{J}_{\text{set}}) = \left\{
\begin{array}{ll}
1 & \text{if } c_i \in \mathcal{J}_{\text{set}}, \\
0 & \text{otherwise.}
\end{array}
\right.
\]

The Intrinsic Score (ISC) is then computed as:
\[
\text{ISC} = \frac{\sum_{i=1}^{m} I(c_i \in \mathcal{J}_{\text{set}})}{m}.
\]

Here, $m$ denotes the cardinality of \( \mathcal{H}_{\text{set}} \).

\subsection{Extrinsic Evaluation}
Extrinsic evaluation assesses the semantic structure of the JSON by evaluating its ability to answer targeted questions. This method avoids direct comparison to the ground truth JSON (\( \mathcal{G}_{\text{t}} \)), which can vary in representation. Instead, we systematically validate the structure by formulating questions that probe specific semantic elements:
In order to form a minimalistic set of questions that validate the structure, we leverage the paths from the root node to each leaf node in the JSON. The number of paths corresponds to the number of leaf nodes. Let $\mathcal{P}$ denote the set of these paths. Formally, we define $\mathcal{P}$ as:
\[
\mathcal{P} = \{ p_i \mid p_i = (n_1, n_2, \ldots, n_{k-1}) \}
\]

Here, $(n_1, n_2, \ldots, n_k)$ represents a sequence of nodes starting from the root node $n_1$ till a leaf node $n_k$ in the JSON structure. For each path $p_i \in \mathcal{P}$, we prompt an LLM (\( \mathcal{M}_{\text{q}} \)) with \( \mathcal{G}_{\text{t}} \) and $p_i$ as inputs, instructing it to generate a single question \( \mathcal{Q}_{\text{i}} \). This question is targeted so that the value at the leaf node of the path is the expected answer \( \mathcal{K}_{\text{e}} \).

An Evaluator LLM (\( \mathcal{M}_{\text{eval}} \)) takes as input the HySem-generated JSON (\( \mathcal{J}_{\text{p}} \)), a question (\( \mathcal{Q}_{\text{i}} \)), and the expected answer (\( \mathcal{K}_{\text{e}} \)). The Evaluator LLM predicts an answer (\( \mathcal{K}_{\text{p}} \)) and compares its prediction with the expected answer (\( \mathcal{K}_{\text{e}} \)), all in a single pass through the LLM.

\textbf{Hypothesis:} \textit{Verifying each leaf node in the JSON is sufficient to confirm the structure's accuracy. If the JSON is incorrect, it will fail to answer questions about these leaf nodes accurately.}

The Evaluator LLM computes the score for the \(i\)-th question-answer pair as follows:

\[
\mathcal{M}_{\text{eval}}(p_i) = \left\{
\begin{array}{ll}
1 & \text{if } \mathcal{K}_{\text{p}} = \mathcal{K}_{\text{e}}, \\
0 & \text{otherwise.}
\end{array}
\right.
\]



The Extrinsic Score (ESC) is computed as:
\[
ESC = \frac{1}{|\mathcal{P}|} \sum_{p_i \in \mathcal{P}} \text{Pred}(p_i).
\]

\section{Results}

We present the results of HySem, measured on 608 testing samples manually annotated from the open-sourced FinTabNet and PubTabNet datasets. The performance of our pipeline was evaluated using both intrinsic and extrinsic metrics, as described in the previous section, ensuring a thorough and well-rounded assessment. The results demonstrate that HySem delivers competitive performance compared to industry-leading models and outperforms popular open-source LLMs while excelling significantly in terms of \textit{token efficiency}.

\subsection{Baselines and Benchmarking}

Our method uses LLaMA-3-Base as the foundational LLM, which we fine-tune to enhance performance for our specific table transformation task. To assess the improvements gained, we benchmarked HySem against popular open-source and proprietary models, including Meta LLaMA-3-8B-Instruct, Microsoft Phi-3-Medium-128K-Instruct, and OpenAI GPT-4o. Table \ref{tab:bench} shows our benchmarking results.

\textbf{LLaMA-3-8B-Instruct:} HySem surpasses LLaMA-3-8B-Instruct by over 25\% in intrinsic accuracy and 3.97\% in extrinsic scores, proving that our task-specific fine-tuning significantly enhances the model's capabilities.

\textbf{Phi-3-Medium-128K-Instruct:} The Phi-3 model displayed weak performance on intrinsic metrics but fared better in extrinsic evaluations, achieving 82.39\%. HySem outperformed Phi-3 in both areas, demonstrating its superiority as a fine-tuned model specialized for this task.

\textbf{GPT-4o:} While GPT-4o demonstrates superior overall accuracy metrics, largely due to extensive training on diverse datasets, particularly involving HTML and XML table formats \citep{sui2024table}, HySem maintains a competitive edge. Although trailing by 2\% in accuracy, HySem offers substantial advantages such as on-premise deployment on commodity hardware, cost-effectiveness, and enhanced data privacy—essential requirements for many enterprises.

\begin{table*}[t]
\centering
\begin{tabular}{l|c|c} 
\toprule
\textbf{Model} & \textbf{Intrinsic Score} & \textbf{Extrinsic Score} \\
\midrule
Meta-Llama-3-8B-Instruct & 66.25 & 84.42 \\
Phi-3-Medium-128K-Instruct & 56.78 & 82.39 \\
GPT-4o & 93.43 & 90.15 \\
\textbf{HySem (Ours)} & 91.12 & 88.39 \\
\bottomrule
\end{tabular}
\caption{Benchmarking Results}
\label{tab:bench}

\end{table*}

\begin{table*}[t]
\centering
\begin{tabular}{l|c} 
\toprule
\textbf{Description} & \textbf{Metric} \\
\midrule
Number of test samples & 608 \\ 
Number of tokens before Token-based Encoding (A) & 366608 \\ 
Number of tokens after Token-based Encoding (B) & 224082 \\
\textbf{Token Efficiency(\%)} & 38.87 \\  

\bottomrule

\end{tabular}
\caption{Token Efficiency Metrics}
\label{tab:opt}

\end{table*}

\subsection{Token Reduction Efficiency}

We evaluated the token reduction efficiency of HySem’s Context Optimizer, which enhances token usage while maintaining semantic accuracy. HySem improves model efficiency, resulting in faster inference times and reduced computational overhead.

\textbf{Token Efficiency} is defined as:

\begin{equation}
\text{Token Efficiency} = \left(1 - \frac{B}{A}\right) \times 100
\end{equation}

Table \ref{tab:opt} shows that HySem achieves over 38\% token efficiency, reflecting a substantial reduction in the token count required for processing, thereby enhancing the LLM inference throughput. The results demonstrate significant benefits gained with using our Token-based encoding method.

\section{Conclusion and Future Work}

We successfully implemented the HySem LLM Pipeline, which generates semantic JSON output from HTML tables. The Context Optimizer significantly improved context utilization, enabling us to process larger tables. Currently, we are piloting our product with a well-known pharmaceutical enterprise that operates multiple production plants and serves a global customer base. This model powers several downstream applications, including analytics from regulatory compliance documents and the automatic creation of these documents. Our future work includes supporting even larger tables that span multiple pages, developing techniques to improve processing speed, and building custom models for other verticals. 



\bibliographystyle{plainnat} 
\bibliography{custom}


\appendix

\section{Appendix}

\subsection{Related Works}
\textbf{LLMs for Tabular Data}

Tables are a fundamental method for presenting structured information across various industrial applications, including regulatory information in the pharmaceutical industry and financial reports for businesses. Some shared characteristics and inherent challenges in tables as mentioned by \citep{fang2024large} include Heterogeneity \citep{borisov2023language}, Sparsity, Context-based interconnection \citep{liu2023goggle}, Lack of prior knowledge \citep{borisov2022deep, borisov2023language}, etc. Recent advancements in Large Language Models \citep{achiam2023gpt, jiang2023mistral, team2024gemma, dubey2024llama} have shown promise in addressing these challenges for various table related tasks. Some table related tasks and techniques include entity matching and data-imputation \citep{10.1145/3654979, korini2023column}, tabular Q\&A\ \citep{chen2024hiqa, wang2024chainoftable, zhang2023reactable, patnaik2024cabinet}, schema augmentation \citep{10.14778/3574245.3574258}, Serialization \citep{yin20acl, yan2024making}, Table Manipulation \citep{zhang2024tablellm}, Table Understanding \citep{lobo2023matching}, Prompt Engineering \citep{zheng2023chainofthought, ziqi-lu-2023-tab}, Table RAG \citep{sundar-heck-2023-ctbls}. Role-play \citep{zhao2023large}.

\textbf{LLMs and HTML}

Recent research has explored various uses of HTML to enhance NLP tasks. \citep{nakano2021webgpt, yao2022webshop} utilize LLMs for Web Navigation tasks involving HTML input formats. \citep{aghajanyan2022htlm} developed the \textbf{HyperText Language Model (HTLM)}, which leverages HTML's structural elements to improve tasks like zero-shot summarization and question answering. Their approach utilizes HTML for structured prompting and template-based guidance but is largely confined to standard NLP tasks and assumes predefined HTML structures, such as \textit{<title>} elements. \citep{gur-etal-2023-understanding} focused on HTML understanding through tasks like semantic classification, description generation, and autonomous web navigation. They identified the \textbf{context window length} as a significant bottleneck, noting that even models supporting longer token sequences struggle with performance degradation when processing larger snippet sizes.

In contrast to the aforementioned works, our approach focuses not merely on HTML, but specifically on the more complex domain of \textbf{HTML tables} particularly Pharmaceutical tables, which present significant challenges due to their hierarchical and relational structure. HTML tables often contain multi-level headers, rowspan and colspan attributes, and intricate relationships between cell elements, making their transformation into a semantic format far more demanding. Our work uniquely tackles this complexity by performing a \textbf{table transformation task}, converting HTML tables into semantic JSON. This transformation is further enhanced by our \textbf{Context Optimizer}, which efficiently reduces token usage while preserving the intricate relationships between table elements. Unlike previous approaches, which focus on predefined HTML structures, our method handles the complexities of tables dynamically, ensuring both structural accuracy and computational efficiency. \textbf{To the best of our knowledge}, this is the first approach to transform complex HTML tables into semantic JSON in a context-length optimized manner.

\textbf{Context Optimization for Tables in LLMs}

Recent advancements in context-length optimization have primarily focused on improving LLM performance for text-based tasks \citep{pan-etal-2024-llmlingua, fountas2024human, han-etal-2024-lm}, but limited work has been done to address the unique challenges presented by tables. \citep{sui2024table} tackled context length limitations by serializing tables into text formats. While these methods help fit table data within the model’s context window, serialization can lead to the loss of structural nuances and detailed information. In contrast, \citep{tian2024spreadsheetllm} introduces a novel encoding method tailored for spreadsheet data. While effective in reducing token usage and computational costs, it addresses a different use-case (spreadsheets) compared to our work (HTML tables). Additionally, the encoding method primarily leverages format and structure based optimisations, whereas our approach rewrites the input to ensure alignment with the tokenizer. Our approach specifically targets HTML tables of arbitrary structure, ensuring that hierarchical and relational elements are preserved during HTML to JSON transformation task. By utilizing our Context Optimizer, we ensure semantic integrity is maintained even during token pruning, unlike text-based methods where reducing detail (e.g., abbreviating "potassium clavulanate" to "potassium") would alter the meaning. This is a significant advantage in transforming HTML tables into semantic JSON without compromising accuracy. 

\subsection{Complex Pharmaceutical Data Table}

\begin{figure}[H]
    \centering
    \begin{minipage}{0.80\textwidth}
        \centering
        \includegraphics[width=0.8\linewidth, height=0.4\textheight]{  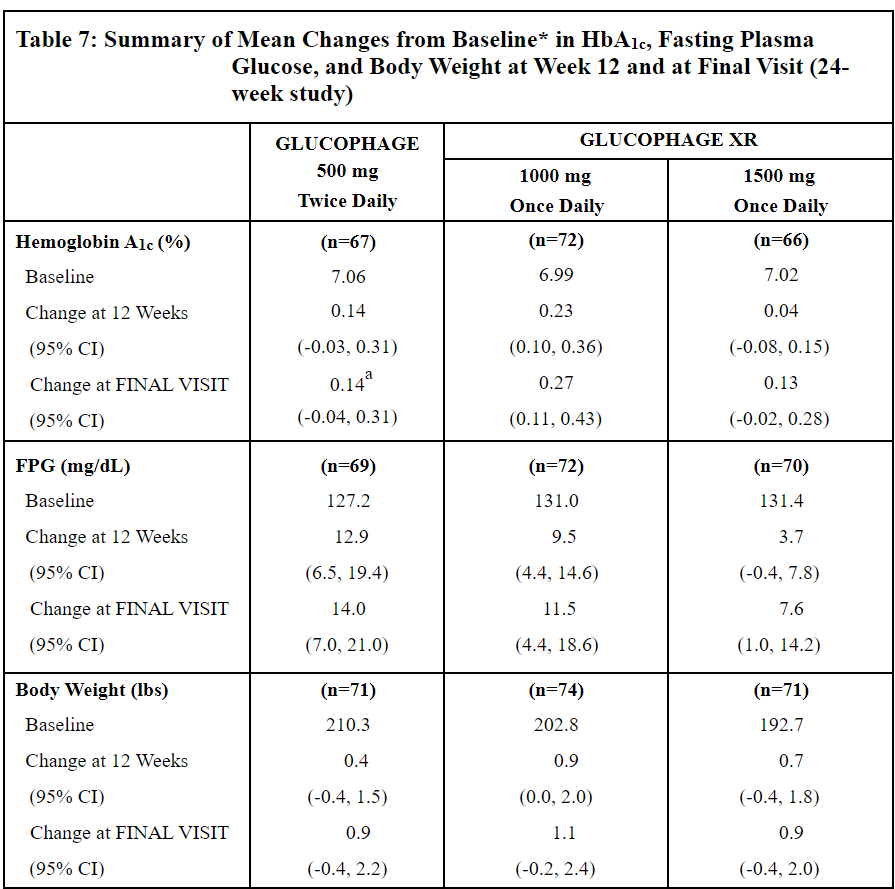}
        \caption{This table compares various dosage regimens of Glucophage and Glucophage XR. The challenge in converting it to semantic JSON lies in its nested categories, including dosage, measurement types, and time points. Each combination of dose, metric, time point, and statistical details (mean change, confidence intervals) must be accurately mapped to database schema keys. Ensuring data integrity while managing variability in column structures is essential for creating a usable semantic representation.}
        \label{fig:first-image}
    \end{minipage}\hfill
    \end{figure}

\subsection{Semantic JSON}

\begin{figure}[H] 
    \centering
    \begin{minipage}{0.32\textwidth}
        \centering
        \includegraphics[width=\linewidth, height=0.6\textheight]{  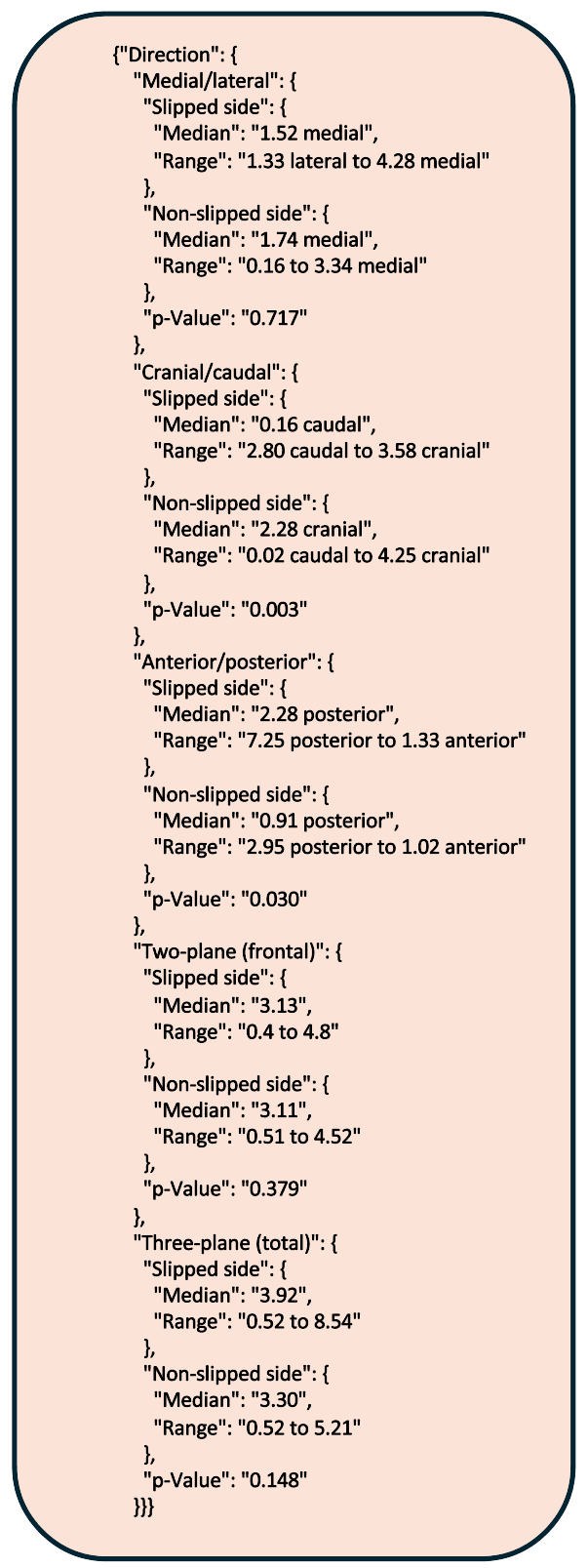}
        \caption{Hysem JSON}
        \label{fig:hysem_json}
    \end{minipage}
    \begin{minipage}{0.32\textwidth}
        \centering
        \includegraphics[width=\linewidth, height=0.6\textheight]{  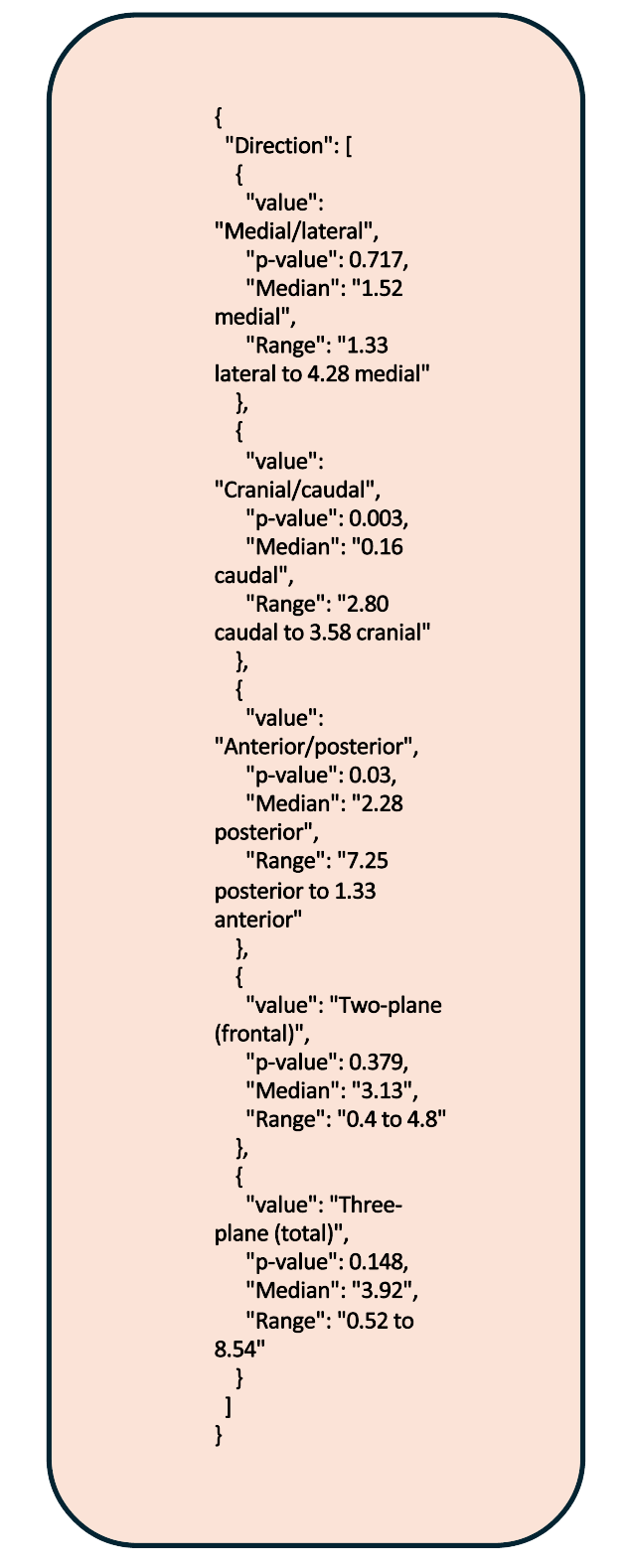}
        \caption{LlaMa3 JSON}
        \label{fig:llama3_json}
    \end{minipage}
    \begin{minipage}{0.32\textwidth}
        \centering
        \includegraphics[width=\linewidth, height=0.6\textheight]{  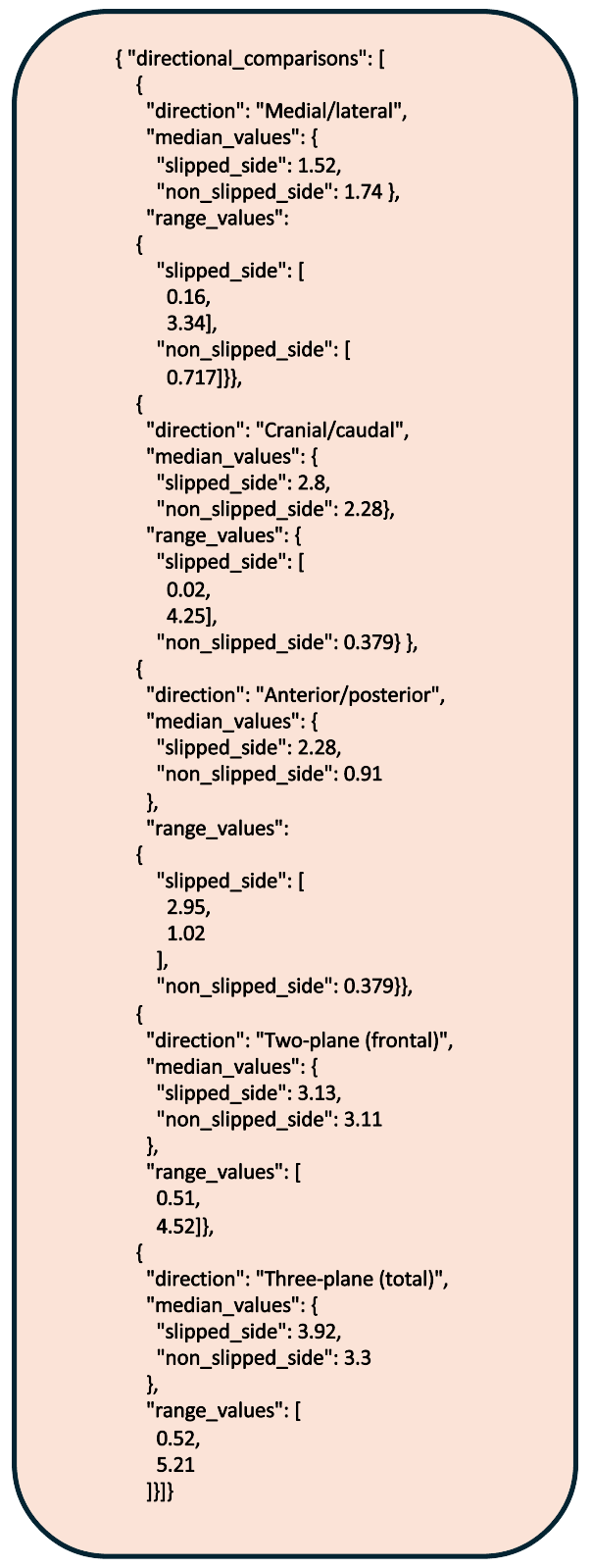}
        \caption{Phi3 JSON}
        \label{fig:phi3_json}
    \end{minipage}
\end{figure}

In our paper, the term \textbf{"Semantic JSON"} refers to a JSON representation where the keys are designed to accurately mirror the hierarchical structure of the JSON tree and align directly with the database schema. This ensures that the JSON output is not only well-structured but also semantically meaningful. Figure \ref{fig:hysem_json} provides an example of such a Semantic JSON output, as generated by Hysem. To highlight the effectiveness of our approach, we compare the outputs generated by other models, namely Llama 3 8B Instruct and Microsoft Phi 3 128K Instruct, as illustrated in Figures \ref{fig:llama3_json} \& \ref{fig:phi3_json} respectively. These illustrations show how the JSON representations from these models differ from ours, particularly in terms of hierarchical accuracy and schema alignment.

\lstset{
  basicstyle=\ttfamily\small,
  columns=flexible,
  breaklines=true,
  numbers=left,
  numberstyle=\tiny\color{gray},
  backgroundcolor=\color{lightgray},
  captionpos=b,
  frame=single,
  rulecolor=\color{black},
  keywordstyle=\color{blue},
  commentstyle=\color{green},
  stringstyle=\color{red},
  showstringspaces=false
}

\begin{figure}[H] 
    \centering
    \begin{minipage}{1.\textwidth}
        \centering
        \includegraphics[width=1.\linewidth, height=0.16\textheight]{ 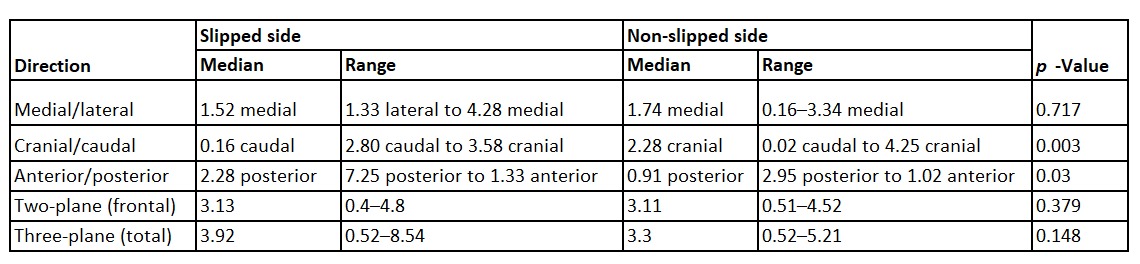}
        \caption{Sample Table to illustrate Semantic JSON}
        \label{fig:appendix_table1}
    \end{minipage}
\end{figure}

\textbf{HySem output versus Meta LLama 3 8B Instruct and Microsoft Phi 3 128K Instruct}

The sample illustrates that the Llama 3 output has missed certain fields, such as 'Slipped Side' and 'Non-slipped Side.' In contrast, Phi 3 has introduced several errors, including misplacing values under the incorrect hierarchy. For instance, the p-value of 0.717 is erroneously placed under the 'range\_values' subtree for the 'Non-slipped Side.'


\subsection{Evaluation Methodology Illustrations}
The performance of HySem is assessed using both intrinsic and extrinsic evaluation methods. Intrinsic evaluation focuses on how accurately the input HTML strings are converted to JSON, while extrinsic evaluation assesses how effectively the semantic structure is preserved.

\begin{figure}[H]
    \centering
    \includegraphics[width=1\linewidth, height=0.12\textheight]{  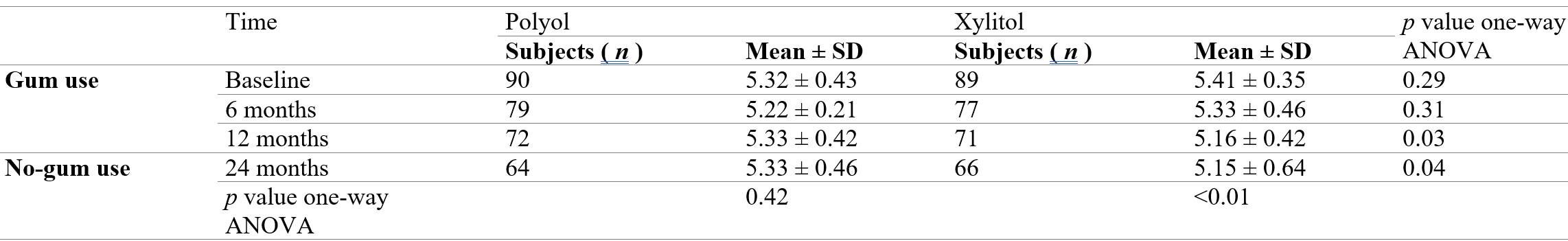}
    \caption{Sample of HTML Table for illustration}
    \label{fig:html_table}
\end{figure}

For the given HTML table, the HySem generated semantic JSON and the ground-truth (GT) JSON are as below.

\begin{figure}[H] 
    \centering
    \begin{minipage}{0.32\textwidth}
        \centering
        \includegraphics[width=\linewidth, height=0.6\textheight]{  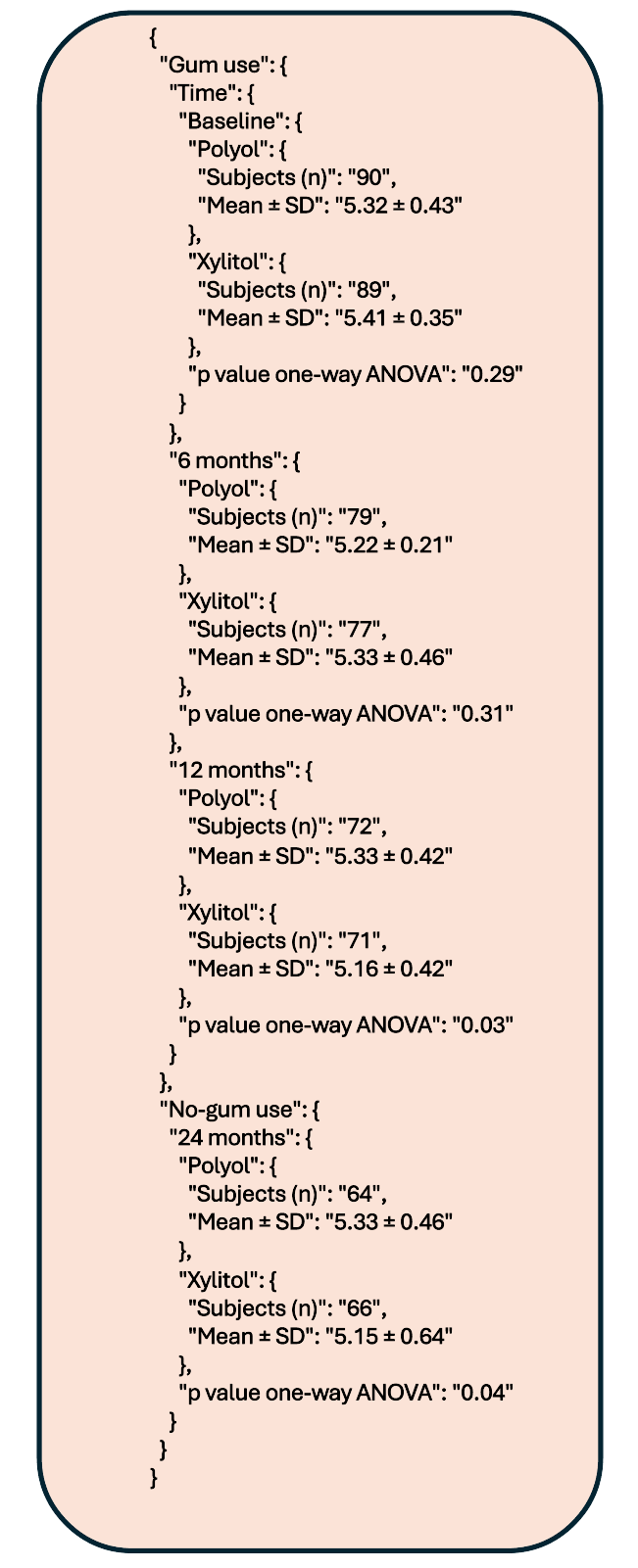}
        \caption{Hysem JSON}
        \label{fig:hysem_eval_json}
    \end{minipage}
    \hfill 
    \begin{minipage}{0.32\textwidth}
        \centering
        \includegraphics[width=\linewidth, height=0.6\textheight]{  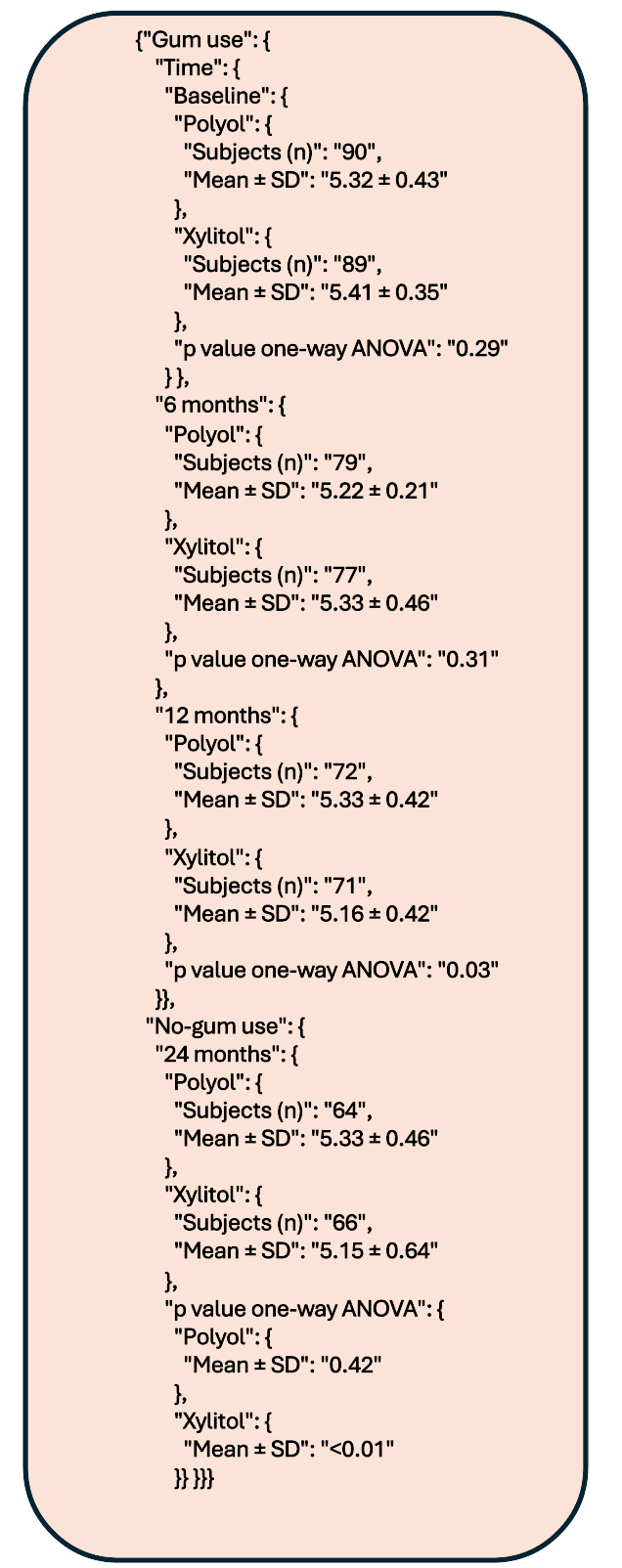}
        \caption{GT JSON}
        \label{fig:gt_json}
    \end{minipage}
\end{figure}

The \textbf{Intrinsic Evaluator's} evaluation results are presented in Table \ref{tab:intrinsic}. Here, each unique cell content of the original HTML table is compared directly to the nodes of the JSON. 

We initially used metrics to measure how well the JSON preserved the \textit{"frequency"} of each string from the HTML. However, this method presented problems. For e.g., a cell content could appear repeatedly in the JSON for each row entry, unlike its single occurrence in the HTML. This discrepancy can skew the accuracy of metrics based on string frequency.

To address this, we simplified the approach by ensuring each unique string from the HTML appears at least once in the JSON. While it’s a reasonable estimate of performance, we recognise that this is an area for future improvement.

\begin{table}[H]
    \centering
    \begin{tabularx}{\textwidth}{X c}  
        \toprule
        \textbf{Cell Content} & \textbf{Presence in Hysem JSON (Yes/No)} \\ 
        \midrule
        "Gum use" & \color{green!80} \checkmark \\
        "Time" & \color{green!80} \checkmark \\
        "Baseline" & \color{green!80} \checkmark \\
        "Polyol" & \color{green!80} \checkmark \\
        "Subjects (n)" & \color{green!80} \checkmark \\
        "90" & \color{green!80} \checkmark \\
        "Mean ± SD" & \color{green!80} \checkmark \\
        "5.32 ± 0.43" & \color{green!80} \checkmark \\
        "Xylitol" & \color{green!80} \checkmark \\
        "89" & \color{green!80} \checkmark \\
        "5.41 ± 0.35" & \color{green!80} \checkmark \\
        "p value one-way ANOVA" & \color{green!80} \checkmark \\
        "0.29" & \color{green!80} \checkmark \\
        "6 months" & \color{green!80} \checkmark \\
        "79" & \color{green!80} \checkmark \\
        "5.22 ± 0.21" & \color{green!80} \checkmark \\
        "77" & \color{green!80} \checkmark \\
        "5.33 ± 0.46" & \color{green!80} \checkmark \\
        "0.31" & \color{green!80} \checkmark \\
        "12 months" & \color{green!80} \checkmark \\
        "72" & \color{green!80} \checkmark \\
        "5.33 ± 0.42" & \color{green!80} \checkmark \\
        "71" & \color{green!80} \checkmark \\
        "5.16 ± 0.42" & \color{green!80} \checkmark \\
        "0.03" & \color{green!80} \checkmark \\
        "No-gum use" & \color{green!80} \checkmark \\
        "24 months" & \color{green!80} \checkmark \\
        "64" & \color{green!80} \checkmark \\
        "5.33 ± 0.46" & \color{green!80} \checkmark \\
        "66" & \color{green!80} \checkmark \\
        "5.15 ± 0.64" & \color{green!80} \checkmark \\
        "0.04" & \color{green!80} \checkmark \\
        "<0.01" & \color{red!80} \textbf{x} \\
        "0.42" & \color{red!80} \textbf{x} \\
        \bottomrule
        \textbf{Intrinsic Score:} & \textbf{94.11\%}
    \end{tabularx}
    \caption{Intrinsic Evaluation results}
    \label{tab:intrinsic}

\end{table}

The \textbf{Extrinsic evaluation} measures the semantic accuracy of the LLM generated JSON. In the example table, there are 21 unique paths from the root node to each leaf node of the GT JSON. As mentioned, the number of paths is equal to the number of leaf nodes. LLM (\( \mathcal{M}_{\text{q}} \)) accepts GT JSON and a path and outputs a single question such that the expected answer is the leaf node of the path. Table \ref{tab:extrinsic2} illustrates the paths, corresponding LLM generated questions and the GT answer.

\begin{table}[H]
\centering
\begin{tabular}{@{}p{4.cm} p{6.cm} p{2.5cm} @{}}
\toprule
\textbf{Path} & \textbf{Question} & \textbf{GT Answer} \\ 
\midrule
Gum use > Time > Baseline > Polyol > Subjects (n) & What is the number of subjects in the baseline group that used gum containing polyol? & 90 \\

Gum use > Time > Baseline > Polyol > Mean ± SD & What is the average amount of gum used by subjects at baseline, with a standard deviation? & 5.32 ± 0.43 \\

Gum use > Time > Baseline > Xylitol > Subjects (n) & What is the number of subjects in the baseline group that used Xylitol? & 89 \\

Gum use > Time > Baseline > Xylitol > Mean ± SD & What is the average amount of Xylitol used by subjects at baseline, with a standard deviation? & 5.41 ± 0.35 \\

Gum use > Time > Baseline > p value one-way ANOVA & What is the p-value of the one-way ANOVA test for the baseline data? & 0.29 \\

Gum use > 6 months > Polyol > Subjects (n) & What is the number of subjects in the group that used gum for 6 months and was given polyol? & 90\\

Gum use > 6 months > Polyol > Mean ± SD & What is the mean value of gum usage after 6 months for polyol? & 5.22 ± 0.21 \\

Gum use > 6 months > Xylitol > Subjects (n) & What is the number of subjects in the group that used Xylitol for 6 months? & 77 \\

Gum use > 6 months > Xylitol > Mean ± SD & What is the mean value of Xylitol at 6 months, along with its standard deviation? & 5.33 ± 0.46 \\

Gum use > 6 months > p value one-way ANOVA & What is the p-value of the one-way ANOVA test for the comparison between Polyol and Xylitol at 6 months? & 0.31 \\

Gum use > 12 months > Polyol > Subjects (n) & How many subjects were in the group that used gum for 12 months and had a polyol treatment? & 72 \\

Gum use > 12 months > Polyol > Mean ± SD & What is the average value of gum use for 12 months with polyol, along with its standard deviation? & 5.33 ± 0.42 \\

Gum use > 12 months > Xylitol > Subjects (n) & What is the number of subjects in the group that used Xylitol for 12 months? & 71 \\

Gum use > 12 months > Xylitol > Mean ± SD & What is the mean value of Xylitol at 12 months, along with its standard deviation? & 5.16 ± 0.42 \\

Gum use > 12 months > p value one-way ANOVA & What is the p-value of the one-way ANOVA test for the comparison at 12 months? & 0.03 \\

No-gum use > 24 months > Polyol > Subjects (n) & How many subjects were in the group that did not use gum and had a follow-up at 24 months, with a focus on polyol? & 64 \\

No-gum use > 24 months > Polyol > Mean ± SD & What is the mean value of gum usage after 24 months for subjects using polyol, along with its standard deviation? & 5.33 ± 0.46 \\

No-gum use > 24 months > Xylitol > Subjects (n) & How many subjects were in the group that did not use gum and had data collected at 24 months, with a focus on xylitol? & 66 \\

No-gum use > 24 months > Xylitol > Mean ± SD & What is the mean value of Xylitol at 24 months for subjects with no gum use, along with its standard deviation? & 5.15 ± 0.64 \\

No-gum use > 24 months > p value one-way ANOVA > Polyol > Mean ± SD & What is the mean difference in standard deviation of 'p value one-way ANOVA' for 'Xylitol' for 'Polyol' at 24 months in a group with no gum use? & 0.42 \\

No-gum use > 24 months > p value one-way ANOVA > Xylitol > Mean ± SD & What is the mean difference in standard deviation of 'p value one-way ANOVA' for 'Xylitol' at '24 months' under 'No-gum use'? & < 0.01 \\

\bottomrule
\end{tabular}
\caption{Paths, Questions and GT Answers of Extrinsic Evaluation}
\label{tab:extrinsic2}
\end{table}

The evaluator LLM (\( \mathcal{M}_{\text{eval}} \)), predicts an answer, given the LLM-generated JSON, a single question, and the expected answer as inputs. It then compares its prediction with the expected answer to generate a score, all in a single pass through the LLM. Table \ref{tab:extrinsic} shows the predicted answers and the extrinsic scores computed for each question, along with the overall extrinsic score.

\begin{table}[H]
\centering
\begin{tabular}{l|c}
\toprule
\textbf{Predicted Answer} & \textbf{Score} \\ 
\midrule
90 & 1 \\
\\
5.32 ± 0.43 & 1 \\
\\

89 & 1 \\
\\

5.41 ± 0.35 & 1 \\
\\

0.29 & 1 \\
\\

90 & 1 \\
\\

5.22 ± 0.21 & 1 \\
\\

77 & 1 \\
\\

5.33 ± 0.46 & 1 \\
\\

0.31 & 1 \\
\\

72 & 1 \\
\\

5.33 ± 0.42 & 1 \\
\\

71 & 1 \\
\\

5.16 ± 0.42 & 1 \\
\\

0.03 & 1 \\
\\

64 & 1 \\
\\

5.33 ± 0.46 & 1 \\
\\

66 & 1 \\
\\

5.15 ± 0.64 & 1 \\
\\

0.04 & 0 \\
\\

0.03 & 0 \\

\bottomrule
\textbf{Extrinsic Score} & 90.47\%

\end{tabular}

\caption{Extrinsic Evaluation results}
\label{tab:extrinsic}

\end{table}

\subsection{Syntax Corrector: Algorithm}

Our Semantic Synthesizer incorporates a custom fine-tuned model specialized in generating JSON outputs. However, occasional syntax errors may occur in the generated output. The Syntax Corrector, part of the HySem pipeline, handles these erroneous JSONs by performing LLM-assisted auto-correction via self-reflection.
We instruct the LLM with the following prompt:

\texttt{"The JSON given below contains syntax errors. Your task is to correct them and provide the corrected output. Provide ONLY the corrected output, without any additional explanation. 'input\_string': \{json\_input\}"}

Algorithm \ref{alg:syntax_corrector} presents the syntax correction algorithm.

\begin{algorithm}[H]
\caption{Syntax Corrector}
\label{alg:syntax_corrector}
\begin{algorithmic}[1]
    \Require Syntactically invalid JSON \Jinv{} from LLM
    \Ensure Syntactically valid JSON \Jval{}
    \State \Jcur{} $\gets$ \Jinv{}
    \State iterations $\gets$ 0
    
    \While{syntax errors in \Jcur{} \textbf{and} iterations $<$ max\_iterations}
        \If{\Jcur{} is syntactically invalid}
            \State \Jcur{} $\gets$ \SCA{}(\Jcur{})
        \EndIf
        \State iterations $\gets$ iterations + 1
    \EndWhile
    
    \State \Jval{} $\gets$ \Jcur{}
    
    \State \textbf{return} \Jval{}
\end{algorithmic}
\end{algorithm}






\end{document}